\documentclass[conference]{ieeeconf}

\IEEEoverridecommandlockouts

\usepackage{cite}
\usepackage{amsmath,amssymb,amsfonts}
\usepackage{algorithmic}
\usepackage{graphicx}
\usepackage{textcomp}
\usepackage{xcolor}
\def\BibTeX{{\rm B\kern-.05em{\sc i\kern-.025em b}\kern-.08em
    T\kern-.1667em\lower.7ex\hbox{E}\kern-.125emX}}
\begin{document}

\title{A Soft Fabric-Based Thermal Haptic Device for VR and Teleoperation}


\author{Rui Chen$^{1*}$, Domenico Chiaradia$^{1}$, Antonio Frisoli$^{1}$,  Daniele Leonardis$^{1}$
\thanks{Domenico Chiaradia was founded by the Next Generation EU project ECS00000017 'Ecosistema dell'Innovazione' Tuscany Health Ecosystem (THE, PNRR, Spoke 9: Robotics and Automation for Health). Rui Chen and Antonio Frisoli are partially funded by the mentioned Next Generation EU project ECS00000017, and partially by MSCA-DN / Project 101073374 - ReWIRE. Views and opinions expressed are however those of the author(s) only and do not necessarily reflect those of the European Union or the European Research Executive Agency (REA). Neither the European Union nor the granting authority can be held responsible for them. }%
\thanks{$^{1}$ The authors are with Institute of Mechanical Intelligence, School of Advanced Studies Sant'Anna(SSSA), 56127 Pisa, Italy, (corresponding email: rui.chen@santannapisa.it)}%
}

\maketitle
\begin{abstract}
This paper presents a novel fabric-based thermal-haptic interface for virtual reality and teleoperation. It integrates pneumatic actuation and conductive fabric with an innovative ultra-lightweight design, achieving only 2~g for each finger unit. By embedding heating elements within textile pneumatic chambers, the system delivers modulated pressure and thermal stimuli to fingerpads through a fully soft, wearable interface.

Comprehensive characterization demonstrates rapid thermal modulation with heating rates up to 3$^{\circ}$C/s, enabling dynamic thermal feedback for virtual or teleoperation interactions. The pneumatic subsystem generates forces up to 8.93~N at 50~kPa, while optimization of fingerpad-actuator clearance enhances cooling efficiency with minimal force reduction. Experimental validation conducted with two different user studies shows high temperature identification accuracy (0.98 overall) across three thermal levels, and significant manipulation improvements in a virtual pick-and-place tasks. Results show enhanced success rates (88.5\% to 96.4\%, p = 0.029) and improved force control precision (p = 0.013) when haptic feedback is enabled, validating the effectiveness of the integrated thermal-haptic approach for advanced human-machine interaction applications.
\end{abstract}


\section{Introduction}

Haptic feedback constitutes a critical component in virtual reality (VR) applications, providing rich sensory information that enables complex task execution and enhances user immersion~\cite{frisoli2024HapticsReview, Tanacar2023VR,bai2021VR}. Modern haptic systems integrate various actuators into wearable devices to deliver comprehensive tactile sensations, including pressure, vibration, deformation, texture, and temperature, effectively simulating real-world interactions~\cite{Pacchierotti2024TeleoperationReview}.

Pneumatic actuators have emerged as a promising solution for haptic feedback delivery, offering significant advantages in weight reduction, inherent compliance, and cost-effectiveness~\cite{sonar2021SPASkin, Adilkhanov2022HapticDevice}. Recent advances in rubber-based pneumatic systems have demonstrated successful sensor integration for closed-loop control~\cite{Choi2023IntegretedSensingSilicone}, while their expansive volume capabilities enable substantial tactile variations~\cite{Cai2024ViboPneumo, qi2023haptglove, Beek2024PUC, Talhan2023HumanTouchFabric}. Within this domain, fabric-based pneumatic actuators have gained particular attention due to their minimal weight and manufacturing simplicity, exemplified by innovations such as pouch motors~\cite{Niiyama2014pouch, niiyama2015pouch}. Notable implementations include WRAP, a four-motor pneumatic system for directional guidance in medical interventions~\cite{Raitor2017WRAPGuidanceFabric}, systems creating continuous lateral motion sensations through coordinated actuation~\cite{Wu2019LateralMotionFabric}, and devices simulating handshake sensations via strategically positioned actuators~\cite{yamaguchi2023handshakepouch}.

Thermal feedback represents another crucial dimension in haptic interfaces, leveraging the skin's dense thermoreceptor network capable of detecting precise temperature variations. This sensory capability provides essential environmental information, driving development of various thermal haptic devices based on different heat transfer mechanisms~\cite{lee2021thermolHapticReview, Raza2024Multi-Modal}.

 Tactile sensory substitution offers potential for symbolic thermal feedback, yet these technologies remain constrained by difficulties in replicating authentic thermal sensations~\cite{visell2009TactileSensorySubstitution, eagleman2023FutureSensorySubstitution}. Fluid-based thermal systems employ temperature-controlled reservoirs with valve-controlled mixing of hot and cold fluids, enabling precise temperature regulation through flow ratio modulation~\cite{Goetz2020PATCHPump}. However, these systems present design complexities and require stable temperature maintenance. Fluid selection involves trade-offs: water-based systems increase weight, while air-based systems exhibit slower thermal response~\cite{Car2020ThermAirGlovePump, Liu2021ThermoCaressMovingPump}.

Thermoelectric (Peltier) devices represent a widely adopted approach, creating temperature differentials through current flow and enabling precise bidirectional temperature control~\cite{shilpa2023PeltierReview}. While successfully integrated with both motorized~\cite{gabardi2018development, Kang2024FlipPeltier,kim2024wirelesslyPeltier, Kang2024FlipPeltier} and pneumatic systems~\cite{Zhang2021PneuModThermal,lee2021three-axis,lee2024softThermolPeltier} to provide rapid responses and broad temperature ranges, they face significant challenges including heat accumulation, weight, rigidity, and system complexity~\cite{lee2021thermolHapticReview, Nimmagadda2021PelterReview}. Recent advances in flexible thermoelectric devices have substantially improved module flexibility, yet limitations remain in the efficiency of these devices ~\cite{kim2020thermalDisplayFlexiblePeltiers,hong2019wearableFlexiblePeltier,Kim2020FlexiblePeltier,nasser2020thermalcaneFlexiblePeltier}.

Motivated by the need for enhanced compactness, compliance, and wearability in thermal haptic interfaces, we present a fully fabric-based thermal and pneumatic haptic system featuring fabric-based thermal haptic actuators (FTHAs). Each FTHA integrates a pneumatic pouch motor with a fabric electric heater, delivering congruent haptic and thermal feedback to fingertips. While currently limited to heating capabilities, the system achieves exceptional thinness, lightness, and compliance with rapid dynamic response characteristics.

This paper presents comprehensive FTHA characterization including thermal step response and force output analysis, alongside user studies evaluating device effectiveness in VR applications. Our findings demonstrate that this novel approach could significantly impact VR or teleoperation interfaces, particularly for applications requiring lightweight, compliant, and responsive thermal haptic feedback.


\section{Soft Thermal Haptic Device}

\begin{figure}[tb]
\centerline{\includegraphics[width=0.5\textwidth]{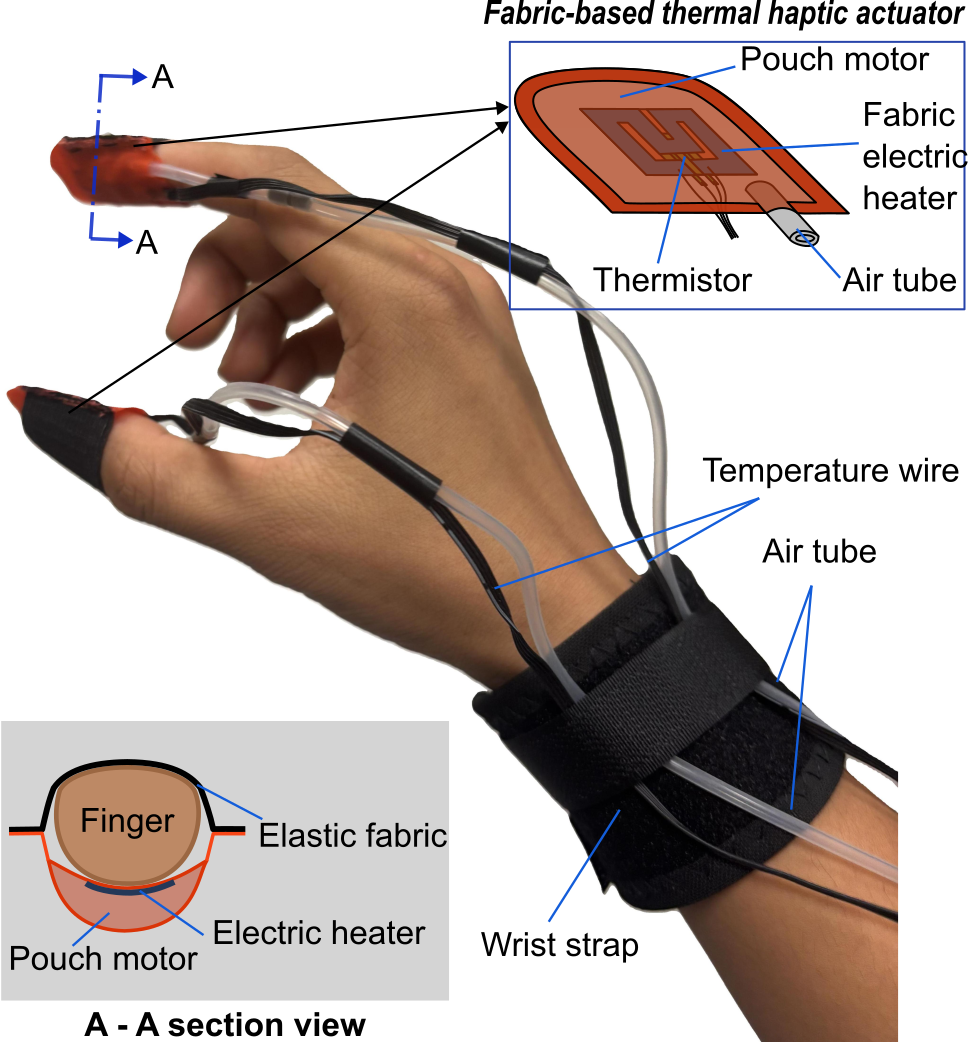}}
\caption{Conceptual illustration of the proposed soft fabric-based thermal haptic device integrating dual-modal feedback capabilities for fingertip interaction.}
\label{Fig.conceptual}
\end{figure}

The proposed soft thermal haptic device integrates two FTHAs within a wearable wrist strap system, worn at the index and thumb fingers, as illustrated in Fig.~\ref{Fig.conceptual}. The FTHAs deliver congruent haptic and thermal feedback to users' fingertips, creating a multimodal tactile experience more informative of the virtual or remote interaction.
The device employs an ergonomic wrist strap design that efficiently manages pneumatic tubes and electrical connections while ensuring stable positioning during use. The FTHAs mount on the index finger and thumb through elastic fabric interfaces, accommodating various finger sizes.

A key innovation is the ultra-lightweight construction, achieving a mass of only 2~g of the FTHA, a substantial reduction compared to conventional electromagnetic motor-based haptic systems. Typically, fingertip haptic devices are in the 10 to 50 g range, depending on the typology of the actuators \cite{pacchierotti2017wearable} and intensity of the feedback. This minimal mass is crucial for prolonged-use applications, such as VR training simulations or therapeutic interventions, substantially reducing user fatigue while maintaining optimal haptic performance~\cite{adilkhanov2022hapticWeight}.

\subsection{FTHA Architecture}

\begin{figure*}[t]
    \centerline{\includegraphics[width=1\textwidth]{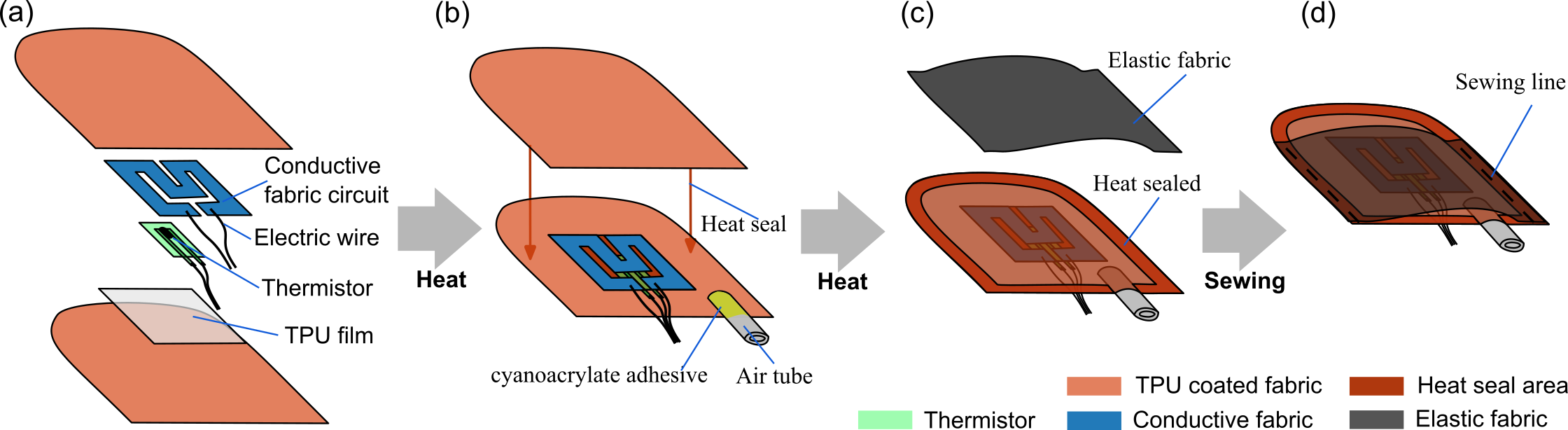}}
    \caption{FTHA fabrication process: (a) Component preparation including TPU-coated fabrics and conductive circuit, (b) Integration of thermistor and conductive fabric circuit, (c) Heat-sealing of pneumatic chamber with air tube attachment, (d) Final assembly with elastic fabric interface.}
    \label{Fig.fabrication}
\end{figure*}

The FTHA integrates two primary functional subsystems: a pneumatic pouch actuator for normal force modulation and an electric heating system for thermal sensation, as detailed in Fig.~\ref{Fig.fabrication}(a). The design follows key guidelines including thermal operation within a comfortable range (25--50$^{\circ}$C) with precise control ($\pm$1$^{\circ}$C), perceptible haptic feedback (more than 1 N force), lightweight construction on fingertip (fabrics with a denier of 40D or less are recommended), and safe temperature limits (50$^{\circ}$C) with proper electrical insulation. 

The pneumatic subsystem utilizes dual-layer fabric sheets with Thermoplastic Polyurethane (TPU) coating. When pressurized, the pouch motor generates normal force against the fingertip while conforming to the finger's natural contour.

The thermal subsystem integrates a fabric-based electric circuit with an NTC thermistor for temperature sensing, placed on the same layer of the conductive fabric. The conductive fabric element, fabricated with precise laser cutting, generates controlled heat when current is applied due to Joule effect. A closed-loop PID control system is implemented to control the desired temperature. While heating is actively controlled, cooling relies on passive mechanisms that include natural convection and thermal dissipation through the device structure.

\subsection{Fabrication Methodology}

The fabrication process follows a systematic approach to ensure consistent device performance and reliability. Precision laser cutting produces the required components: two rectangular TPU-coated fabric sections (Adventure Experts, Slovenia, 40-denier) and a square woven conductive fabric circuit. A supplementary TPU film facilitates component integration, as shown in Fig.~\ref{Fig.fabrication}(a).

Assembly begins with precise positioning and thermal bonding of the NTC thermistor (50~k$\Omega$ at 25$^{\circ}$C) and fabric circuit onto one TPU-coated fabric layer using the prepared TPU film (Fig.~\ref{Fig.fabrication}(b)). The second TPU-coated fabric layer is then heat-sealed to the circuit-integrated layer, forming the pneumatic pouch motor structure. A TPU air tube is bonded using cyanoacrylate adhesive to enable pneumatic control (Fig.~\ref{Fig.fabrication}(c)). Final assembly involves the sewing of an elastic fabric interface to create an ergonomic fingertip attachment (Fig.~\ref{Fig.fabrication}(d)).

Critical design parameters were optimized for thermal and haptic performance. The conductive circuit dimensions (15~mm $\times$ 15~mm) provide adequate fingertip contact area, with 3~mm-wide circuit traces ensuring uniform current distribution. The thermistor's central placement allows accurate temperature measurement of the active area. Pouch actuators dimensions were chosen based on average finger anthropometrics: 18~mm (breadth) $\times$ 16~mm for the index finger and 20~mm (breadth) $\times$ 18~mm for the thumb, ensuring reliable contact across diverse user populations \cite{hsiao2015Finger}.

\subsection{Thermal Response Characterization}

\begin{figure}[t]
\centerline{\includegraphics[width=0.5\textwidth]{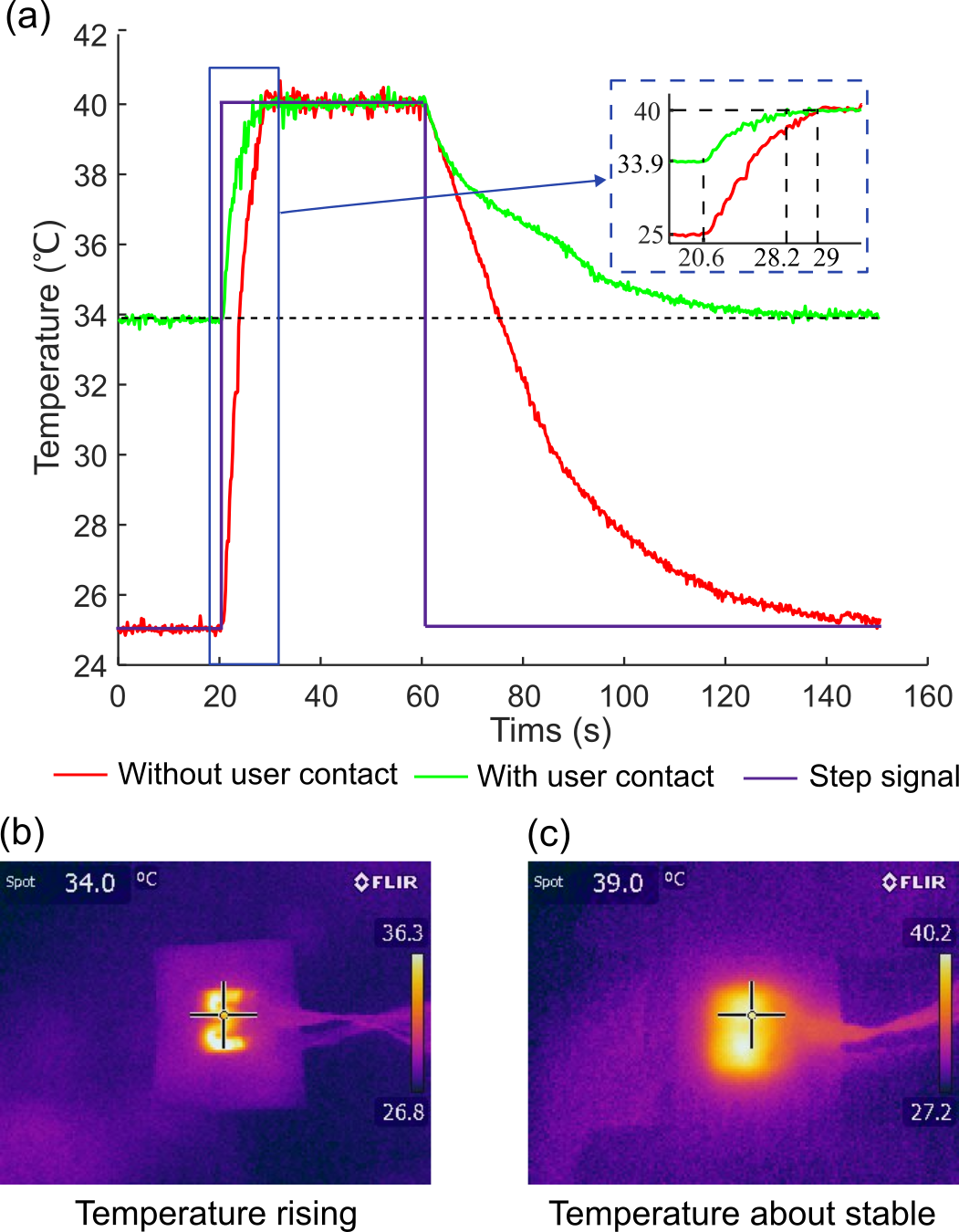}}
\caption{Thermal characterization of FTHAs: (a) Temperature step response under unloaded and finger-contact conditions, (b) Thermal image during temperature rise showing distinct coil heating pattern, (c) Thermal image near steady state showing diffused heat distribution.}
\label{Fig.temperature}
\end{figure}

A comprehensive thermal characterization study was conducted to evaluate the device's temperature control capabilities under both unloaded and finger-contact conditions. The experimental protocol employed a series of step input signals from 25$^{\circ}$C to 40$^{\circ}$C, each lasting 40 seconds. Temperature data was acquired at 5~Hz sampling rate, enabling detailed analysis of thermal dynamics as illustrated in Fig.~\ref{Fig.temperature}(a).

Under unloaded conditions, the actuator exhibited rapid initial heating with a peak heating rate of approximately 3$^{\circ}$C/s during the initial phase, attributed to maximum temperature differential. The system achieved the target temperature of 40$^{\circ}$C within 8.4~seconds, demonstrating an average heating rate of 1.79$^{\circ}$C/s. The cooling phase exhibited notably different dynamics, requiring approximately 90~seconds to return to baseline temperature with a maximum cooling rate of 0.4$^{\circ}$C/s during the initial cooling period. Thermal imaging revealed distinct coil patterns during temperature rise due to large gradients between the heated coil and surrounding fabric (Fig.~\ref{Fig.temperature}(b)), which became diffused as temperature approached steady state (Fig.~\ref{Fig.temperature}(c)).

Under finger-contact conditions, thermal behavior demonstrated significant modifications due to bio-thermal interactions. The baseline temperature stabilized at approximately 34$^{\circ}$C, reflecting thermal equilibrium between the actuator and finger tissue. The heating phase from 34$^{\circ}$C to 40$^{\circ}$C required 7.6~seconds, achieving an average heating rate of 0.79$^{\circ}$C/s. The cooling phase exhibited significantly extended duration, requiring approximately 70~seconds to return to the finger-contact baseline temperature—a substantial difference compared to unloaded conditions.

This observed disparity in cooling rates between contact and non-contact conditions reveals a critical design consideration for thermal haptic systems. The enhanced cooling efficiency in unloaded states suggests that the strategic implementation of a controlled clearance between finger and actuator during deflated states could significantly improve thermal responsiveness. This finding indicates that optimizing the actuator-fingerpad gap when thermal feedback is not required could enhance overall system performance by facilitating more rapid thermal state transitions, while maintaining effective force transmission during inflated states.

\subsection{Force Characterization and Clearance Optimization}

\begin{figure}[htp]
\centerline{\includegraphics[width=0.5\textwidth]{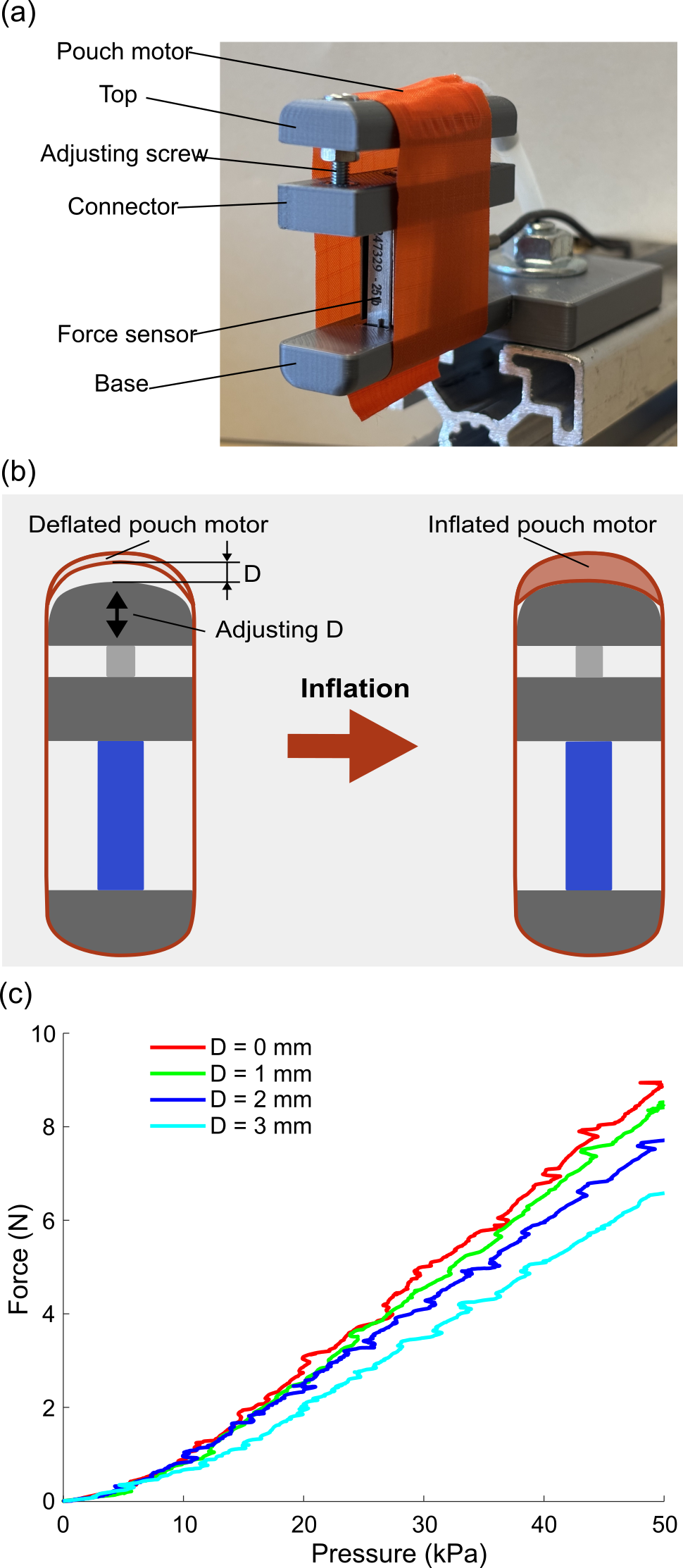}}
\caption{Force characterization of FTHAs: (a) Experimental setup with anatomically-curved testing platform, (b) Schematic diagram showing clearance control mechanism, (c) Force output versus input pressure under different clearance conditions.}
\label{Fig.force}
\end{figure}

Building upon thermal characterization findings that demonstrated enhanced cooling rates with increased actuator-finger clearance, a comprehensive force characterization study was conducted to quantify the trade-off between thermal performance and force output capabilities. A specialized experimental apparatus was developed to systematically evaluate FTHA force output characteristics under varying clearance conditions, as illustrated in Fig.~\ref{Fig.force}(a) and (b).

The testing platform comprises five key components: an anatomically-curved top plate, dual precision adjustment screws, connector assembly, high-precision force sensor, and contoured base plate. The curved profiles of both plates were engineered to replicate geometric constraints of human fingertip interaction, enabling accurate simulation of force transfer dynamics during actual device operation. The dual adjustment screws facilitate precise clearance control between the deflated actuator and top plate, enabling systematic investigation of clearance effects on force output characteristics.

The experimental protocol began with precise calibration of the initial contact state. Adjustment screws were carefully manipulated until a minimal preloading force of 0.05~N was measured by the force sensor, establishing the zero-clearance (D = 0~mm) reference position. Subsequent force measurements were conducted by incrementally increasing the input pressure from 0 to 50~kPa, while simultaneously recording pressure and force data. This procedure was systematically repeated across multiple clearance settings (D = 0, 1, 2, and 3~mm) to characterize the relationship between clearance, input pressure, and output force.

The experimental results, presented in Fig.~\ref{Fig.force}(c), demonstrate a predominantly linear relationship between input pressure and output force across all tested clearance values. Maximum force output of 8.93~N was achieved at 50~kPa with D = 0~mm. Systematic but moderate decreases in maximum force output were observed with increasing clearance values. Specifically, clearances of 1~mm, 2~mm, and 3~mm yielded maximum force outputs of approximately 8.5~N (5\% reduction), 7.7~N (14\% reduction), and 6.6~N (26\% reduction), respectively.



\section{User Study}

\begin{figure*}[h!]
\centerline{\includegraphics[width=1\textwidth]{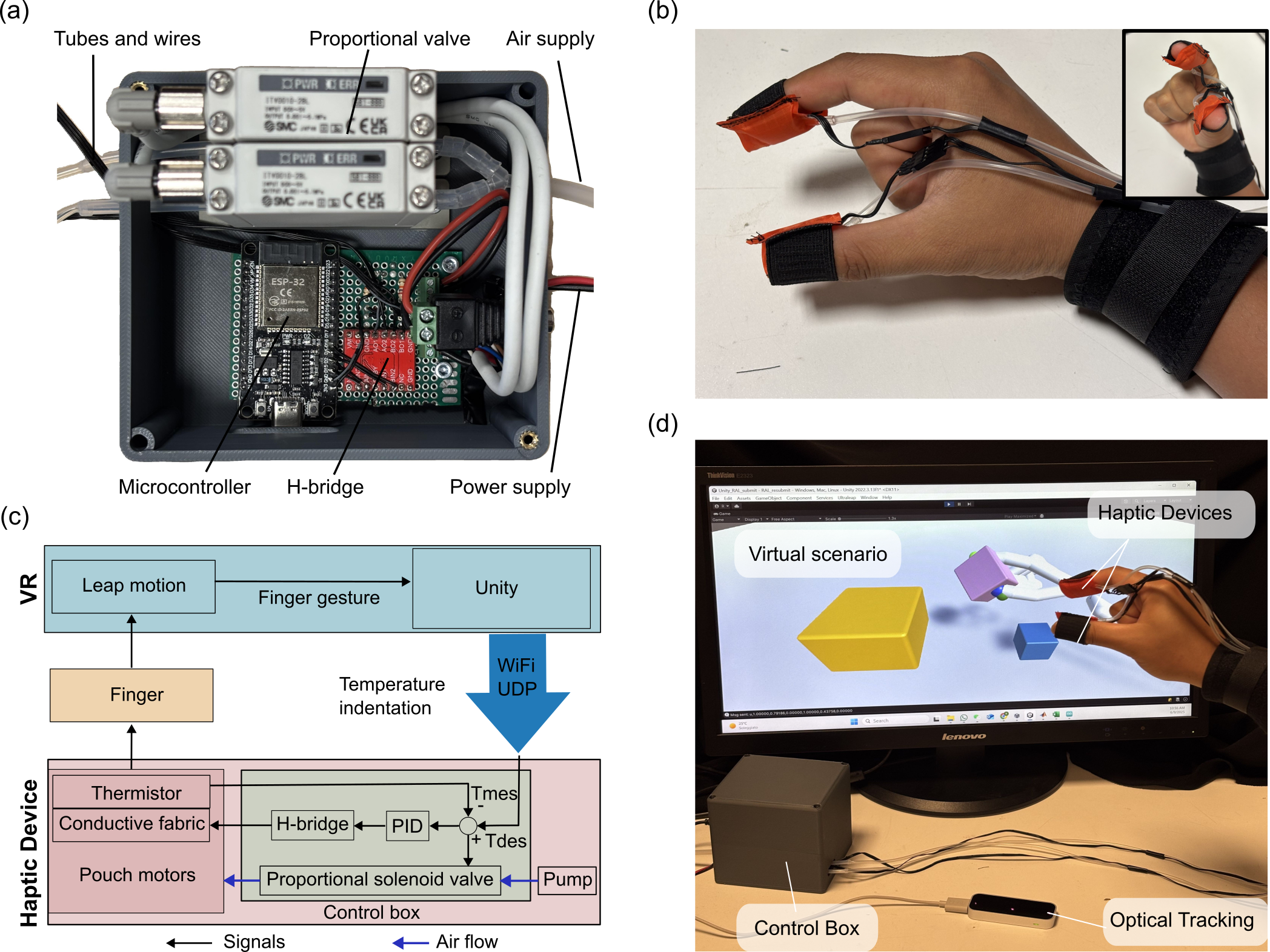}}
\caption{Experimental setup: (a) Control box components, (b) Thermal feedback protocol setup, (c) Haptic feedback system schematic showing VR integration, (d) Virtual manipulation experimental environment.}
\label{Fig.setup}
\end{figure*}

\subsection{System Overview}

Two user studies were conducted to investigate thermal haptic device performance. A preliminary study combining thermal and haptic feedback is detailed in supplementary materials (Figs.~S1-S2). Then, the main experimental evaluation separated thermal feedback (temperature discrimination) and haptic feedback (virtual manipulation) protocols as described below.

The control system (Fig.~\ref{Fig.setup}(a)) features an ESP-32 microcontroller serving as the central hub. Two proportional solenoid valves (ITV0010, SMC) regulate air supply to the pouch motors, achieving precise pressure and force modulation. Fabric heater operation is managed through current control using an H-bridge driver IC. Temperature feedback utilizes NTC thermistor resistance measurements in a closed-loop PID controller to maintain desired temperatures.

Eleven healthy subjects participated to the experiments (7 male, 4 female; aged 28.8~$\pm$~3.1 years), all right-handed. The study was approved from the Ethical Board of the Scuola Superiore Sant'Anna (protocol 412023) with informed consent obtained from all participants.

\subsection{Thermal Feedback Experiment}

Three temperature levels were evaluated: cool (25$^{\circ}$C, heater off), warm (40--41$^{\circ}$C), and hot (43--44$^{\circ}$C). Subjects distinguished temperatures using fingertip sensation exclusively, as shown in Fig.~\ref{Fig.setup}(b).

A 5-minute familiarization period was performed to present all the temperature stimuli to the subject. Then, stimuli were presented in randomized order (18 trials in total, 6 per stimulus). After each stimulus, participants had to answer which of the three temperatures they perceived. To ensure consistent fingertip-actuator contact, actuators maintained 10~kPa pressure during testing. Between trials, actuators were deflated for approximately 20~seconds to enable natural cooling. Response times were recorded for each trial, encompassing heating, identification, and response recording phases.

\subsection{Haptic Feedback Experiment}

The VR-integrated haptic system (Fig.~\ref{Fig.setup}(c)) employed vision-based hand tracking (Leap Motion, Ultraleap) to capture hand gestures and map them to a virtual hand representation. Virtual hand interactions with objects generated corresponding indentation data transmitted to the haptic device via WiFi using UDP protocol.

The interaction mechanism used fingertip coordinates from the tracking device to control two proxy points in VR. These proxies were connected via virtual springs to virtual spheres (green spheres at fingertips) featuring collisions and physical interaction as rigid bodies in the virtual physics simulation. The implemented virtual spring method~\cite{leonardis20163} enabled modulated object grasping while maintaining stability of the physics simulation, especially during the grasping condition.

In the experimental environment (Fig.~\ref{Fig.setup}(d)) a cube (purple object) had to be picked from a base (yellow stand) and precisely placed on a smaller target area (blue stand on the side). The user could interact with the cube only through the collision-enabled spheres at the index and thumb fingertips. During contact, indentation was calculated as the distance between the tracking proxy and the corresponding collision sphere. The virtual interaction force was proportional to the indentation multiplied by the virtual stiffness coefficient.

\begin{figure*}[t]
\centerline{\includegraphics[width=1\textwidth]{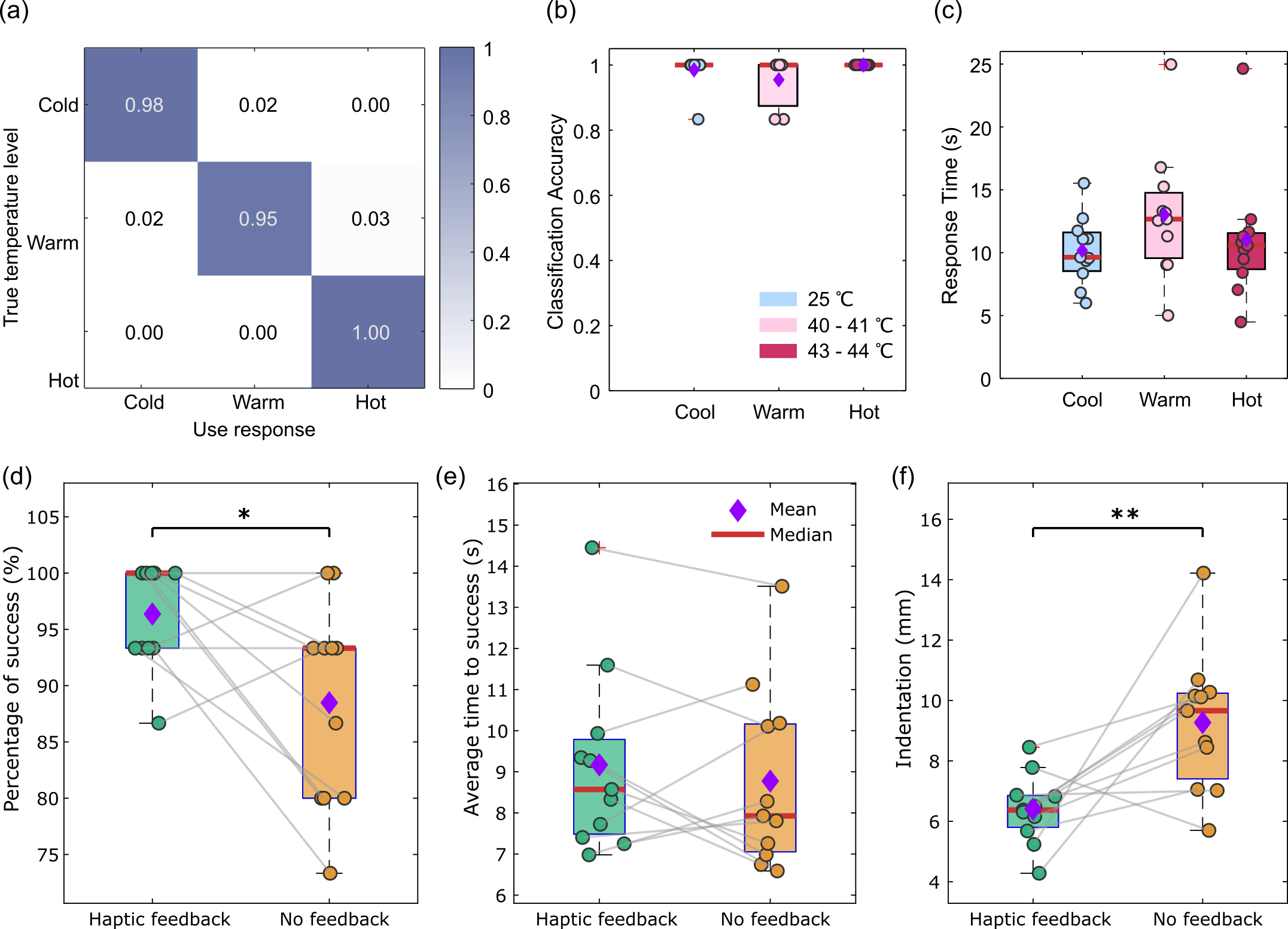}}
\caption{User study results: (a) Temperature identification confusion matrix, (b) Temperature classification accuracy by category, (c) Response time distribution for temperature levels, (d) Task success rate comparison, (e) Average completion time comparison, (f) Finger indentation comparison between haptic conditions.}
\label{Fig.subjects}
\end{figure*}

Two experimental conditions were evaluated: (1) Haptic Feedback (HF)—subjects wore the haptic device receiving both visual and tactile feedback; (2) No Feedback (NF)—subjects removed the haptic device and manipulated objects with visual feedback only. Pneumatic pressure was proportional to indentation up to 20~kPa at 20~mm maximum indentation. For subjects with smaller fingers, additional soft fabric padding was placed between finger and elastic fabric to ensure proper actuator contact.

Subjects performed object grasping and placement tasks on targets for 15 trials per condition. Successful placement required maintaining object stability on the target for 1~seconds. Object falls resulted in failed attempts. Condition order was randomized to minimize learning effects (See supplementary video).

Data collection encompassed task completion time, per-trial duration, failed attempts, and finger indentations. Measured metrics included success rate (success trials divided by total trials), average time to success (total time divided by success trials), and average indentation (mean index and thumb indentation when fingers were colliding with the cube). Statistical analysis employed paired t-test between conditions.

\section{Experimental Results}

\subsection{Thermal Feedback Performance}

Temperature identification achieved high overall accuracy of 0.98 across all conditions, as shown in Fig.~\ref{Fig.subjects}(a) and (b). Individual temperature accuracies were 0.98 (cool), 0.95 (warm), and 1.0 (hot). Median accuracy reached 1.0 across all temperature levels despite occasional classification errors in cool and warm conditions.

Response times varied by temperature level: mean values were 10.2~s (cool), 13~s (warm), and 11~s (hot), with an overall mean of 11.41~s (Fig.~\ref{Fig.subjects}(c)). Response times included heating duration, subject identification period, and response recording time. The slightly longer response time for warm temperatures may reflect the intermediate nature of this thermal stimulus requiring more deliberative discrimination.

\subsection{Haptic Feedback Performance}

Virtual manipulation results (Fig.~\ref{Fig.subjects}(d)-(f)) demonstrated significant benefits of haptic feedback on task performance. The averaged success rates increased from 88.5\% (NF) to 96.4\% (HF), with statistical significance (p = 0.029), indicating that haptic feedback substantially enhances task completion reliability.

Task completion times showed no significant difference between conditions: averaged times to success were 8.8~s (NF) versus 9.2~s (HF) (p = 0.455). This can be explained by the fact  that while haptic feedback provides additional informative sensory cues, hence improving grasping modulation and success rate, such information needs to be processed by the user, limiting or counter-balancing its impact on the execution time.

Finger indentation analysis revealed significant behavioral differences (p = 0.01). The haptic feedback condition resulted in reduced average indentations of 6.4~mm compared to 9.3~mm without haptic feedback. This finding indicates that haptic feedback enables more controlled and gentle object manipulation, suggesting improved force modulation and spatial awareness during virtual object interaction.

\section{Discussion}

In this work we presented a fully fabric-based approach to design a wearable fingertip haptic device. By integrating thermal features directly within the pneumatic structure, our approach differs from other pneumatic wearable devices by enabling congruent multimodal feedback with reduced complexity and faster response compared to other pneumatic or fluid-based systems~\cite{Goetz2020PATCHPump,Car2020ThermAirGlovePump}. The fully fabric construction achieves ultra-lightweight actuation at only 2~g per module, substantially lighter than haptic devices usign conventional electromagnetic motors ~\cite{Gabardi2018ThermolModulePeltier,Kang2024FlipPeltier}, while providing enhanced wearability and mobility compared to desktop thermal interfaces~\cite{lee2021three-axis}. The soft fabric heaters overcome the inherent rigidity limitations of conventional Peltier-based thermal systems~\cite{lee2024softThermolPeltier, Zhang2021PneuModThermal} and demonstrate superior flexibility (please see supplementary video) compared to flexible thermoelectric devices~\cite{kim2020thermalDisplayFlexiblePeltiers,hong2019wearableFlexiblePeltier,Kim2020FlexiblePeltier,nasser2020thermalcaneFlexiblePeltier}.

The characterization studies reveal a critical trade-off between thermal and force performance in finger-actuator clearance design. Enhanced cooling rates observed when the finger is not in contact with the actuator suggest that increased fingerpad-actuator clearance could accelerate cooling during deflated states. However, increased clearance correspondingly reduces force output capabilities. Despite the challenge of precise clearance adjustment, our analysis indicates that a 2~mm clearance represents an optimal compromise, maintaining 86\% of maximum force output while significantly improving thermal responsiveness.

Two user studies validated the device's functionality through complementary approaches. A preliminary study integrated thermal and haptic feedback within a VR manipulation task (detailed in supplementary materials), while separate studies evaluated temperature identification and virtual manipulation to validate individual modality performance. These studies demonstrate the device's effectiveness across different interaction paradigms. It has to be noted that in the experimental comparison the No Haptic condition was performed with bare-hands, differently from the conventional procedure of just disabling the haptic device. Hence, the improved performance in the Haptic condition already includes any possible effect on dexterity that the wearing of the haptic interface can involve.

Subtle differences between study protocols yield insights into system optimization. The virtual manipulation protocol comparison reveals that proportional pressure control (matching indentation to pressure) achieves superior force regulation compared to binary on-off control. This is evidenced by greater and more significant indentation reduction with haptic feedback under proportional control (Fig.~\ref{Fig.subjects}(f)) versus on-off control (Fig.~S2(a), supplementary text), indicating more precise force modulation capabilities.

Temperature feedback evaluation demonstrates context-dependent performance characteristics. In isolated temperature identification tasks, subjects exhibited deliberative behavior with extended response times (11.41~s overall) but achieved high accuracy (0.98 overall). Conversely, the preliminary experiment (included in the supplementary material) with integrated thermal-haptic VR manipulation showed that thermal discrimination, conducted along the pick and place task, carried a minimal temporal overhead (3.06~s increase compared to haptic-only conditions), suggesting effective multimodal integration.

Several limitations warrant future investigation. First, passive cooling limitations may impact user experience during extended VR sessions; active chamber venting represents a viable enhancement strategy. Second, observed latency between VR interaction and pneumatic actuation (visible in supplementary video) requires systematic characterization and optimization. Third, while conductive fabric offers flexibility and lightweight advantages, alternative materials including copper foil, liquid metal circuits, or conductive yarns merit investigation for enhanced thermal performance. Finally, additional rendering techniques (i.e. vibration modulation) can provide additional tactile dimensions or alternative force rendering approaches, necessitating comparative evaluation in similar fine manipulation scenarios.

\section{Conclusions}

This study presents a novel lightweight fabric-based thermal haptic device for human-machine interaction, envisaging applications such as teleoperation and virtual reality interactions. The device integrates fabric-based electric heating and pneumatic actuation to deliver congruent thermal and haptic feedback to fingertips. Weighing only 2~g per thimble, the system achieves force output up to 8.93~N at 50~kPa with rapid thermal response (3$^{\circ}$C/s peak heating rate).

Characterization studies establish optimal design parameters, identifying 2~mm clearance as the ideal compromise maintaining 86\% of maximum force while enhancing thermal responsiveness. User studies validate device effectiveness: temperature identification achieved 98\% accuracy, while virtual manipulation showed significant improvements in task success rates (88.5\% to 96.4\%, p = 0.029) and force control precision (p = 0.01). A pilot experiment showed similar benefits with a combined thermal-haptic feedback delivered in the same manipulation task. 
Noticeably, improved performance were achieved versus a bare-hands No Feedback condition, highlighting the wearability and preservation of user's dexterity of the proposed approach.
This ultra-lightweight design establishes a foundation for extended-use VR and teleoperation applications including training simulations, therapeutic interventions, and immersive entertainment.
\section*{Acknowledgment}


\bibliographystyle{IEEEtran}
\bibliography{mybib} 

\begin{thebibliography}{10}
\providecommand{\url}[1]{#1}
\csname url@samestyle\endcsname
\providecommand{\newblock}{\relax}
\providecommand{\bibinfo}[2]{#2}
\providecommand{\BIBentrySTDinterwordspacing}{\spaceskip=0pt\relax}
\providecommand{\BIBentryALTinterwordstretchfactor}{4}
\providecommand{\BIBentryALTinterwordspacing}{\spaceskip=\fontdimen2\font plus
\BIBentryALTinterwordstretchfactor\fontdimen3\font minus \fontdimen4\font\relax}
\providecommand{\BIBforeignlanguage}[2]{{%
\expandafter\ifx\csname l@#1\endcsname\relax
\typeout{** WARNING: IEEEtran.bst: No hyphenation pattern has been}%
\typeout{** loaded for the language `#1'. Using the pattern for}%
\typeout{** the default language instead.}%
\else
\language=\csname l@#1\endcsname
\fi
#2}}
\providecommand{\BIBdecl}{\relax}
\BIBdecl

\bibitem{frisoli2024HapticsReview}
A.~Frisoli and D.~Leonardis, ``Wearable haptics for virtual reality and beyond,'' \emph{Nature Reviews Electrical Engineering}, pp. 1--14, 2024.

\bibitem{Tanacar2023VR}
N.~T. Tanacar, M.~H. Mughrabi, A.~U. Batmaz, D.~Leonardis, and M.~Sarac, ``The impact of haptic feedback during sudden, rapid virtual interactions,'' in \emph{2023 IEEE World Haptics Conference (WHC)}, 2023, pp. 64--70.

\bibitem{bai2021VR}
H.~Bai, S.~Li, and R.~F. Shepherd, ``Elastomeric haptic devices for virtual and augmented reality,'' \emph{Advanced Functional Materials}, vol.~31, no.~39, p. 2009364, 2021.

\bibitem{Pacchierotti2024TeleoperationReview}
C.~Pacchierotti and D.~Prattichizzo, ``Cutaneous/tactile haptic feedback in robotic teleoperation: Motivation, survey, and perspectives,'' \emph{IEEE Transactions on Robotics}, vol.~40, pp. 978--998, 2024.

\bibitem{sonar2021SPASkin}
H.~A. Sonar, J.-L. Huang, and J.~Paik, ``Soft touch using soft pneumatic actuator--skin as a wearable haptic feedback device,'' \emph{Advanced Intelligent Systems}, vol.~3, no.~3, p. 2000168, 2021.

\bibitem{Adilkhanov2022HapticDevice}
A.~Adilkhanov, M.~Rubagotti, and Z.~Kappassov, ``Haptic devices: Wearability-based taxonomy and literature review,'' \emph{IEEE Access}, vol.~10, pp. 91\,923--91\,947, 2022.

\bibitem{Choi2023IntegretedSensingSilicone}
H.~Choi, M.~R. Cutkosky, and A.~A. Stanley, ``Integrated pneumatic sensing and actuation for soft haptic devices,'' \emph{IEEE Robotics and Automation Letters}, vol.~8, no.~11, pp. 7591--7598, 2023.

\bibitem{Cai2024ViboPneumo}
S.~Cai, Z.~Chen, H.~Gao, Y.~Huang, Q.~Zhang, X.~Yu, and K.~Zhu, ``Vibopneumo: A vibratory-pneumatic finger-worn haptic device for altering perceived texture roughness in mixed reality,'' \emph{IEEE Transactions on Visualization and Computer Graphics}, pp. 1--14, 2024.

\bibitem{qi2023haptglove}
J.~Qi, F.~Gao, G.~Sun, J.~C. Yeo, and C.~T. Lim, ``Haptglove—untethered pneumatic glove for multimode haptic feedback in reality--virtuality continuum,'' \emph{Advanced Science}, vol.~10, no.~25, p. 2301044, 2023.

\bibitem{Beek2024PUC}
F.~E.~v. Beek, Q.~P.~I. Bisschop, and I.~A. Kuling, ``Validation of a soft pneumatic unit cell (puc) in a vr experience: A comparison between vibrotactile and soft pneumatic haptic feedback,'' \emph{IEEE Transactions on Haptics}, vol.~17, no.~2, pp. 191--201, 2024.

\bibitem{Talhan2023HumanTouchFabric}
A.~Talhan, Y.~Yoo, and J.~R. Cooperstock, ``Soft pneumatic haptic wearable to create the illusion of human touch,'' \emph{IEEE Transactions on Haptics}, vol.~17, no.~2, pp. 177--190, 2024.

\bibitem{Niiyama2014pouch}
R.~Niiyama, D.~Rus, and S.~Kim, ``Pouch motors: Printable/inflatable soft actuators for robotics,'' in \emph{2014 IEEE International Conference on Robotics and Automation (ICRA)}, 2014, pp. 6332--6337.

\bibitem{niiyama2015pouch}
R.~Niiyama, X.~Sun, C.~Sung, B.~An, D.~Rus, and S.~Kim, ``Pouch motors: Printable soft actuators integrated with computational design,'' \emph{Soft Robotics}, vol.~2, no.~2, pp. 59--70, 2015.

\bibitem{Raitor2017WRAPGuidanceFabric}
M.~Raitor, J.~M. Walker, A.~M. Okamura, and H.~Culbertson, ``Wrap: Wearable, restricted-aperture pneumatics for haptic guidance,'' in \emph{2017 IEEE International Conference on Robotics and Automation (ICRA)}, 2017, pp. 427--432.

\bibitem{Wu2019LateralMotionFabric}
W.~Wu and H.~Culbertson, ``Wearable haptic pneumatic device for creating the illusion of lateral motion on the arm,'' in \emph{2019 IEEE World Haptics Conference (WHC)}, 2019, pp. 193--198.

\bibitem{yamaguchi2023handshakepouch}
S.~Yamaguchi, T.~Hiraki, H.~Ishizuka, and N.~Miki, ``Handshake feedback in a haptic glove using pouch actuators,'' in \emph{Actuators}, vol.~12, no.~2.\hskip 1em plus 0.5em minus 0.4em\relax MDPI, 2023, p.~51.

\bibitem{lee2021thermolHapticReview}
J.~Lee, D.~Kim, H.~Sul, and S.~H. Ko, ``Thermo-haptic materials and devices for wearable virtual and augmented reality,'' \emph{Advanced Functional Materials}, vol.~31, no.~39, p. 2007376, 2021.

\bibitem{Raza2024Multi-Modal}
A.~Raza, W.~Hassan, and S.~Jeon, ``Pneumatically controlled wearable tactile actuator for multi-modal haptic feedback,'' \emph{IEEE Access}, vol.~12, pp. 59\,485--59\,499, 2024.

\bibitem{visell2009TactileSensorySubstitution}
Y.~Visell, ``Tactile sensory substitution: Models for enaction in hci,'' \emph{Interacting with Computers}, vol.~21, no. 1-2, pp. 38--53, 2009.

\bibitem{eagleman2023FutureSensorySubstitution}
D.~M. Eagleman and M.~V. Perrotta, ``The future of sensory substitution, addition, and expansion via haptic devices,'' \emph{Frontiers in Human Neuroscience}, vol.~16, p. 1055546, 2023.

\bibitem{Goetz2020PATCHPump}
D.~T. Goetz, D.~K. Owusu-Antwi, and H.~Culbertson, ``Patch: Pump-actuated thermal compression haptics,'' in \emph{2020 IEEE Haptics Symposium (HAPTICS)}, 2020, pp. 643--649.

\bibitem{Car2020ThermAirGlovePump}
S.~Cai, P.~Ke, T.~Narumi, and K.~Zhu, ``Thermairglove: A pneumatic glove for thermal perception and material identification in virtual reality,'' in \emph{2020 IEEE Conference on Virtual Reality and 3D User Interfaces (VR)}, 2020, pp. 248--257.

\bibitem{Liu2021ThermoCaressMovingPump}
\BIBentryALTinterwordspacing
Y.~Liu, S.~Nishikawa, Y.~a. Seong, R.~Niiyama, and Y.~Kuniyoshi, ``Thermocaress: A wearable haptic device with illusory moving thermal stimulation,'' in \emph{Proceedings of the 2021 CHI Conference on Human Factors in Computing Systems}, ser. CHI '21.\hskip 1em plus 0.5em minus 0.4em\relax New York, NY, USA: Association for Computing Machinery, 2021. [Online]. Available: \url{https://doi.org/10.1145/3411764.3445777}
\BIBentrySTDinterwordspacing

\bibitem{shilpa2023PeltierReview}
M.~Shilpa, M.~A. Raheman, A.~Aabid, M.~Baig, R.~Veeresha, and N.~Kudva, ``A systematic review of thermoelectric peltier devices: Applications and limitations,'' \emph{FDMP-Fluid Dynamics \& Materials Processing}, vol.~19, no.~1, pp. 187--206, 2023.

\bibitem{gabardi2018development}
M.~Gabardi, D.~Leonardis, M.~Solazzi, and A.~Frisoli, ``Development of a miniaturized thermal module designed for integration in a wearable haptic device,'' in \emph{2018 IEEE Haptics Symposium (HAPTICS)}.\hskip 1em plus 0.5em minus 0.4em\relax IEEE, 2018, pp. 100--105.

\bibitem{Kang2024FlipPeltier}
\BIBentryALTinterwordspacing
S.~Kang, G.~Kim, S.~Hwang, J.~Park, A.~I. A.~M. Elsharkawy, and S.~Kim, ``Flip-pelt: Motor-driven peltier elements for rapid thermal stimulation and congruent pressure feedback in virtual reality,'' in \emph{Proceedings of the 37th Annual ACM Symposium on User Interface Software and Technology}, ser. UIST '24.\hskip 1em plus 0.5em minus 0.4em\relax New York, NY, USA: Association for Computing Machinery, 2024. [Online]. Available: \url{https://doi.org/10.1145/3654777.3676363}
\BIBentrySTDinterwordspacing

\bibitem{kim2024wirelesslyPeltier}
J.-H. Kim, A.~V{\'a}zquez-Guardado, H.~Luan, J.-T. Kim, D.~S. Yang, H.~Zhang, J.-K. Chang, S.~Yoo, C.~Park, Y.~Wei \emph{et~al.}, ``A wirelessly programmable, skin-integrated thermo-haptic stimulator system for virtual reality,'' \emph{Proceedings of the National Academy of Sciences}, vol. 121, no.~22, p. e2404007121, 2024.

\bibitem{Zhang2021PneuModThermal}
\BIBentryALTinterwordspacing
B.~Zhang and M.~Sra, ``Pneumod: A modular haptic device with localized pressure and thermal feedback,'' in \emph{Proceedings of the 27th ACM Symposium on Virtual Reality Software and Technology}, ser. VRST '21.\hskip 1em plus 0.5em minus 0.4em\relax New York, NY, USA: Association for Computing Machinery, 2021. [Online]. Available: \url{https://doi.org/10.1145/3489849.3489857}
\BIBentrySTDinterwordspacing

\bibitem{lee2021three-axis}
E.-H. Lee, S.-H. Kim, and K.-S. Yun, ``Three-axis pneumatic haptic display for the mechanical and thermal stimulation of a human finger pad,'' in \emph{Actuators}, vol.~10, no.~3.\hskip 1em plus 0.5em minus 0.4em\relax MDPI, 2021, p.~60.

\bibitem{lee2024softThermolPeltier}
S.~Lee, S.~Jang, and Y.~Cha, ``Soft wearable thermo+ touch haptic interface for virtual reality,'' \emph{iScience}, vol.~27, no.~12, 2024.

\bibitem{Nimmagadda2021PelterReview}
L.~A. Nimmagadda, R.~Mahmud, and S.~Sinha, ``Materials and devices for on-chip and off-chip peltier cooling: A review,'' \emph{IEEE Transactions on Components, Packaging and Manufacturing Technology}, vol.~11, no.~8, pp. 1267--1281, 2021.

\bibitem{kim2020thermalDisplayFlexiblePeltiers}
S.-W. Kim, S.~H. Kim, C.~S. Kim, K.~Yi, J.-S. Kim, B.~J. Cho, and Y.~Cha, ``Thermal display glove for interacting with virtual reality,'' \emph{Scientific reports}, vol.~10, no.~1, p. 11403, 2020.

\bibitem{hong2019wearableFlexiblePeltier}
S.~Hong, Y.~Gu, J.~K. Seo, J.~Wang, P.~Liu, Y.~S. Meng, S.~Xu, and R.~Chen, ``Wearable thermoelectrics for personalized thermoregulation,'' \emph{Science advances}, vol.~5, no.~5, p. eaaw0536, 2019.

\bibitem{Kim2020FlexiblePeltier}
\BIBentryALTinterwordspacing
S.~Kim, T.~Kim, C.~S. Kim, H.~Choi, Y.~J. Kim, G.~S. Lee, O.~Oh, and B.~J. Cho, ``Two-dimensional thermal haptic module based on a flexible thermoelectric device,'' \emph{Soft Robotics}, vol.~7, no.~6, pp. 736--742, 2020, pMID: 32286158. [Online]. Available: \url{https://doi.org/10.1089/soro.2019.0158}
\BIBentrySTDinterwordspacing

\bibitem{nasser2020thermalcaneFlexiblePeltier}
A.~Nasser, K.-N. Keng, and K.~Zhu, ``Thermalcane: Exploring thermotactile directional cues on cane-grip for non-visual navigation,'' in \emph{Proceedings of the 22nd international ACM SIGACCESS conference on computers and accessibility}, 2020, pp. 1--12.

\bibitem{pacchierotti2017wearable}
C.~Pacchierotti, S.~Sinclair, M.~Solazzi, A.~Frisoli, V.~Hayward, and D.~Prattichizzo, ``Wearable haptic systems for the fingertip and the hand: taxonomy, review, and perspectives,'' \emph{IEEE transactions on haptics}, vol.~10, no.~4, pp. 580--600, 2017.

\bibitem{adilkhanov2022hapticWeight}
A.~Adilkhanov, M.~Rubagotti, and Z.~Kappassov, ``Haptic devices: Wearability-based taxonomy and literature review,'' \emph{IEEE Access}, vol.~10, pp. 91\,923--91\,947, 2022.

\bibitem{hsiao2015Finger}
H.~Hsiao, J.~Whitestone, T.-Y. Kau, and B.~Hildreth, ``Firefighter hand anthropometry and structural glove sizing: a new perspective,'' \emph{Human factors}, vol.~57, no.~8, pp. 1359--1377, 2015.

\bibitem{leonardis20163}
D.~Leonardis, M.~Solazzi, I.~Bortone, and A.~Frisoli, ``A 3-rsr haptic wearable device for rendering fingertip contact forces,'' \emph{IEEE transactions on haptics}, vol.~10, no.~3, pp. 305--316, 2016.

\bibitem{Gabardi2018ThermolModulePeltier}
M.~Gabardi, D.~Leonardis, M.~Solazzi, and A.~Frisoli, ``Development of a miniaturized thermal module designed for integration in a wearable haptic device,'' in \emph{2018 IEEE Haptics Symposium (HAPTICS)}, 2018, pp. 100--105.

\end{thebibliography}

\end{document}